\documentclass{article} 
\usepackage{nips15submit_e,times}
\usepackage{url}

\title{C3: Lightweight Incrementalized MCMC for\\Probabilistic Programs using\\Continuations and Callsite Caching}

\author{ \hspace{1em} Daniel Ritchie \hspace{2em} Andreas Stuhlm\"{u}ller \hspace{2em} Noah D. Goodman \\\\ Stanford University }

\nipsfinalcopy 

\usepackage{amsmath,listings,xspace,color,amssymb}
\usepackage{amsfonts}
\usepackage{dsfont}
\usepackage{enumitem}
\usepackage{mathtools}
\usepackage{bm}

\usepackage{inconsolata}

\usepackage{etoolbox}
\usepackage{breqn}

\usepackage[export]{adjustbox}

\usepackage{algorithm}
\usepackage[noend]{algpseudocode}

\usepackage{amsthm}

\usepackage{subcaption}

\BeforeBeginEnvironment{dmath}{\begin{nolinenumbers}}
\AfterEndEnvironment{dmath}{\end{nolinenumbers}}
\BeforeBeginEnvironment{dmath*}{\begin{nolinenumbers}}
\AfterEndEnvironment{dmath*}{\end{nolinenumbers}}

\usepackage{booktabs}
\usepackage{multirow}
\usepackage{array}

\usepackage[percent]{overpic}

\usepackage{wrapfig}

\newcommand{\ic}[1]{\lstinline[basicstyle=\fontsize{8pt}{8.25pt}\selectfont\ttfamily]{#1}}

\input{js.def}

\begin{document}

\maketitle

\begin{abstract}
Lightweight, source-to-source transformation approaches to implementing MCMC for probabilistic programming languages are popular for their simplicity, support of existing deterministic code, and ability to execute on existing fast runtimes~\cite{Lightweight}. However, they are also slow, requiring a complete re-execution of the program on every Metropolis Hastings proposal. We present a new extension to the lightweight approach, C3, which enables efficient, incrementalized re-execution of MH proposals. C3 is based on two core ideas: transforming probabilistic programs into continuation passing style (CPS), and caching the results of function calls. We show that on several common models, C3 reduces proposal runtime by 20-100x, in some cases reducing runtime complexity from linear in model size to constant. We also demonstrate nearly an order of magnitude speedup on a complex inverse procedural modeling application.
\end{abstract}

\section{Introduction}
\label{sec:intro}

Probabilistic programming languages (PPLs) are a powerful, general-purpose tool for developing probabilistic models. A PPL is a programming language augmented with random sampling statements; programs written in a PPL correspond to generative priors. Performing inference on such programs amounts to reasoning about the space of execution traces which satisfy some condition on the program output. Many different PPL systems have been proposed, such as BLOG~\cite{BLOG}, Figaro~\cite{Figaro}, Church~\cite{Church}, Venture~\cite{Venture}, Anglican~\cite{Anglican}, and Stan~\cite{Stan}.

There are many possible implementations of PPL inference. One popular choice is the `Lightweight MH' framework~\cite{Lightweight}. Lightweight MH uses a source-to-source transformation to turn a probablistic program into a deterministic one, where random choices are uniquely identified by their structural position in the program execution trace. Random choice values are then stored in a database indexed by these structural `addresses.' To perform a Metropolis-Hastings proposal, Lightweight MH changes the value of a random choice and re-executes the program, looking up the values of other random choices in the database to reuse them when possible. Lightweight MH is simple to implement and allows PPLs to be built atop existing deterministic languages. Users can thus leverage existing libraries and fast compilers/runtimes for these `host' languages. For example, Stochastic Matlab can access Matlab's rich matrix and image manipulation routines~\cite{Lightweight}, WebPPL runs on Google's highly-optimized V8 Javascript engine~\cite{WebPPL}, and Quicksand's host language compiles to fast machine code using LLVM~\cite{Quicksand}.

Unfortunately, Lightweight MH is also inefficient: when an MH proposal changes a random choice, the entire program re-executes to propagate this change. This is rarely necessary: for many models, most proposals affect only a small subset of the program execution trace.
To update the trace, re-execution is needed only where values can change. Under Lightweight MH, random choice values are preserved and reused when possible, limiting the effect of a proposal to a subset of the changed variable's Markov blanket (sometimes a much smaller subset, due to context-specific independence~\cite{CSI}).
Custom PPL interpreters can leverage this property to incrementalize proposal re-execution~\cite{Venture}, but implementing such interpreters is complicated, and using them makes it difficult or impossible to leverage libraries and fast runtimes for existing deterministic languages.

In this paper, we present a new implementation technique for MH proposals on probabilistic programs that gives the best of both worlds: incrementalized proposal execution using a lightweight, source-to-source transformation framework. Our method, C3, is based on two core ideas:
\begin{enumerate}
	\item \emph{Continuations}: Converting the program into continuation-passing style to allow program re-execution to begin anywhere.
	\item \emph{Callsite caching}: Caching function calls to avoid re-execution when function inputs or ouputs have not changed.
\end{enumerate}
We first describe how to implement C3 in any functional PPL with first-class functions; our implementation is integrated into the open-source WebPPL probabilistic programming language~\cite{WebPPL}. We then compare C3 to Lightweight MH, showing that it gives orders of magnitude speedups on common models such as HMMs, topic models, Gaussian mixtures, and hierarchical linear regression. In some cases, C3 reduces runtimes from linear in model size to constant. We also demonstrate that C3 is nearly an order of magnitude faster on a complex inverse procedural modeling example from computer graphics.

\section{Approach}
\label{sec:approach}

\begin{figure}[t!]
\begin{subfigure}[b]{0.48\linewidth}
\begin{lstlisting}
// Hidden Markov Model
var hmm = function(n, obs) {
  if (n === 0)
    return true;
  else {
    var prev = hmm(n-1, obs);
    var state = transition(prev);
    observation(state, obs[n]);
    return state;
  }
};
\end{lstlisting}
\label{fig:motivatingExample_code}
\end{subfigure}
\begin{subfigure}[b]{0.52\linewidth}
\includegraphics[width=\linewidth]{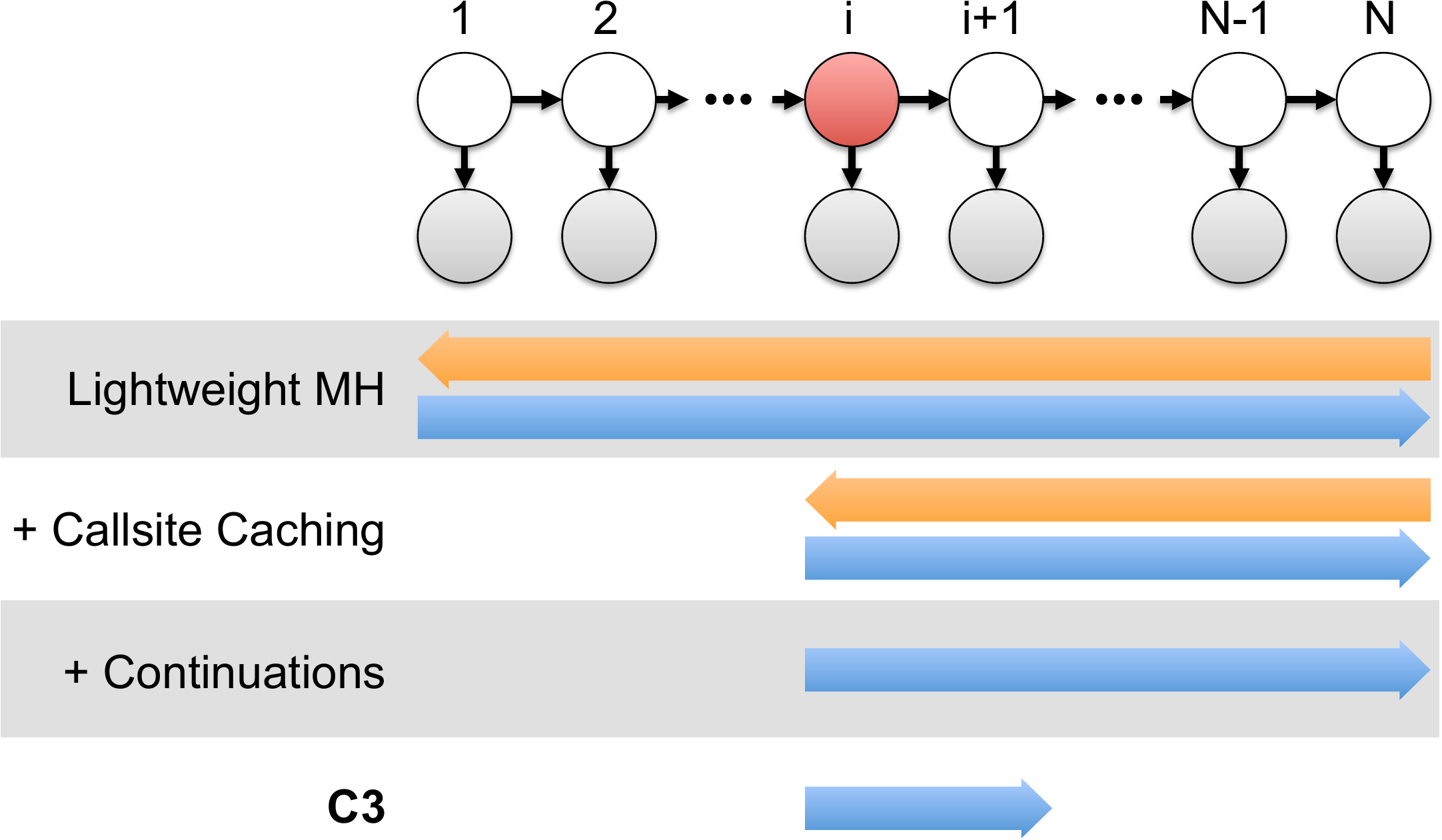}
\label{fig:motivatingExample_illustration}
\end{subfigure}
\caption{\emph{(Left)} A simple HMM program in the WebPPL language. \emph{(Right)} Illustrating the re-execution behavior of different MH implementations in response to a proposal to the random choice $c_i$ shaded in red. Lightweight MH re-executes the entire \ic{hmm} program, invoking (orange bar) and then unwinding (blue bar) the full chain of recursive calls. Callsite caching allows re-execution to skip all recursive calls under \ic{hmm(i-1, obs)}. With continuations, re-execution only has to unwind from the continuation of choice $c_i$. Combining callsite caching and continuations allows re-execution to terminate upon returning from \ic{hmm(i+1, obs)}, since its return value does not change.}
\label{fig:motivatingExample}
\end{figure}

To illustrate our approach, we use a simple example: a binary state Hidden Markov Model program written in WebPPL (Figure~\ref{fig:motivatingExample} Left). This program recursively samples latent states (inside the \ic{transition} function), conditioning on the observations in the \ic{obs} list (inside the \ic{observation} function). When invoked, \ic{hmm(N, obs)} generates a linear chain of latent and observed random variables (Figure~\ref{fig:motivatingExample} Right).

Consider how Lightweight MH performs a proposal on this program. It first runs the program once to initialize the database of random choices.
It then selects a choice $c_i$ uniformly at random from this database (the red circle in Figure~\ref{fig:motivatingExample} Right) and changes its value. This change necessitates a constant-time update to the score of $c_{i+1}$. However, Lightweight MH re-executes the \emph{entire} program, 
invoking a chain of recursive calls to \ic{hmm} (the orange bar in Figure~\ref{fig:motivatingExample} Right) and then unwinding those calls (the blue bar). This process requires $2N$ such call visits for an HMM with $N$ states.

One strategy for speeding up re-execution is to cache function calls and reuse their results if they are invoked again with unchanged inputs. We call this scheme, which is a generalization of Lightweight MH's random choice reuse policy, \emph{callsite caching}. With this strategy, the recursive re-execution of \ic{hmm} must still traverse all ancestors of choice $c_i$ but can stop at \ic{hmm(i, obs)}: it can reuse the result of \ic{hmm(i-1, obs)}, since the inputs have not changed. As shown in Figure~\ref{fig:motivatingExample} Right, using callsite caching can result in less re-execution, but it still requires $\sim 2N$ \ic{hmm} call visits on average.

Now suppose we instead convert the program into continuation passing style. CPS re-organizes a program to make all data and control flow explicit---instead of returning, functions invoke a `continuation' function which represents the remaining computation to be performed~\cite{ContinuationsBook}. For our HMM example, by storing the continuation at $c_i$, computation can resume from the point where this random choice is made, which corresponds to unwinding the stack from \ic{hmm(i, obs)} up to \ic{hmm(N, obs)}. Looking at the `Continuations' row of Figure~\ref{fig:motivatingExample}, this is a significant improvement over Lightweight MH and is also better than callsite caching. However, it still requires $\sim N$ call visits.

Our main insight is that we can achieve the desired runtime by combining callsite caching with continuations---we call the resulting system \textbf{C3}. With C3, re-execution can not only jump directly to choice $c_i$ by invoking its continuation, but it can actually terminate almost immediately: the cache also contains the return values of all function calls, and since the return value of \ic{hmm(i+1, obs)} has not changed, all subsequent computation will not change either. C3 unwinds only two recursive \ic{hmm} calls, giving the desired constant-time update. Thus C3 is more than the sum of its parts: by combining caching with CPS, it enables incrementalization benefits that neither component can deliver independently.

In the sections that follow, we describe how to implement C3 in a functional PPL. Specifically, we describe how to transform the program source at compile-time (Section~\ref{sec:compileTime}) to make requisite data available to the runtime caching mechanism (Section~\ref{sec:runTime}).
\section{Compile-time Source Transformations}
\label{sec:compileTime}

\begin{figure}[t!]

\lstdefinestyle{smallnonums}{numbers=none,basicstyle=\fontsize{6pt}{6.25pt}\selectfont\ttfamily}
\lstset{escapeinside={<@}{@>}}

\begin{minipage}{0.33\linewidth}
\begin{lstlisting}[style=smallnonums]
// Initial HMM code
var hmm = function(n, obs) {
  if (n === 0)
    return true;
  else {
    var prev = hmm(n-1, obs);
    var state = transition(prev);
    observation(state, obs[n]);
    return state;
  }
};
\end{lstlisting}
\end{minipage}
\hspace{-2em}
\begin{minipage}{0.33\linewidth}
\begin{lstlisting}[style=smallnonums]
// After caching transform
var hmm = function(n, obs) {
  if (n === 0)
    return true;
  else {
    var prev = <@\textcolor{purple}{cache(hmm, n-1, obs);}@>
    var state = <@\textcolor{purple}{cache(transition, prev);}@>
    <@\textcolor{purple}{cache(observation, state, obs[n]);}@>
    return state;
  }
};
\end{lstlisting}
\end{minipage}
\begin{minipage}{0.33\linewidth}
\begin{lstlisting}[style=smallnonums]
// After function tagging transform
var hmm = <@\textcolor{purple}{tag(}@>function(n, obs) {
  if (n === 0)
    return true;
  else {
    var prev = cache(hmm, n-1, obs);
    var state = cache(transition, prev);
    cache(observation, state, obs[n]);
    return state;
  }
}<@\textcolor{purple}{, '1', [hmm, transition, observation]);}@>
\end{lstlisting}
\end{minipage}

\caption{Source code transformations used by C3. \emph{(Left)} Original HMM code. \emph{(Middle)} Code after applying the caching transform, wrapping all callsites with the \ic{cache} intrinsic. \emph{(Right)} Code after applying the function tagging transform, where all functions are annotated with a lexically-unique ID and the values of their free variables. An example CPS-transformed program can be found in the ancillary materials.}
\label{fig:sourceTransforms}
\end{figure}

Lightweight MH transforms the source code of probabilistic programs to compute random choice addresses; the transformed code can then be executed on existing runtimes for the host deterministic language. C3 fits into this framework by adding three additonal source transformations: caching, function tagging, and a standard continuation passing style transform for functional languages.

\paragraph{Caching} This transform wraps every function callsite with a call to an intrinsic \ic{cache} function (Figure~\ref{fig:sourceTransforms} Middle). This function performs run-time callsite cache lookups, as described in Section~\ref{sec:runTime}.

\paragraph{Function tagging} This transform analyzes the body of each function and tags the function with both a lexically-unique ID as well as the values of its free variables (Figure~\ref{fig:sourceTransforms} Right). In Section~\ref{sec:runTime}, we describe how C3 uses this information to decide whether a function call must be re-executed.

The final source transformation pipeline is: caching $\rightarrow$ function tagging $\rightarrow$ address computation $\rightarrow$ CPS. Standard compiler optimizations such as inlining, constant folding, and common subexpression elimination can then be applied. In fact, the host language compiler often already performs such optimizations, which is an additional benefit of the lightweight transformational approach.

\section{Runtime Caching Implementation}
\label{sec:runTime}

\begin{figure}[t!]

\lstdefinestyle{smaller}{basicstyle=\fontsize{7pt}{7.25pt}\selectfont\ttfamily}

\begin{minipage}{0.5\linewidth}
\begin{minipage}{\linewidth}
\begin{lstlisting}[style=smaller]
// Arguments added by compiler:
// a: current address
// k: current continuation
function cache(a, k, fn, args) {
  // Global function call stack
  var currNode = nodeStack.top();
  var node = find(a, currNode.children);
  if (node === null) {
    node = FunctionNode(a);
    // Insert maintains execution order
    insert(node, currNode.children,
           currNode.nextChildIndex);
  }
  execute(node, k, fn, args);
}
\end{lstlisting}
\end{minipage}

\begin{minipage}{\linewidth}
\begin{lstlisting}[style=smaller]
// rc: a random choice node
function propagate(rc) {
  // Restore call stack up to rc.parent
  restore(nodeStack, rc.parent);
  // Changes to rc may make siblings unreachable
  markUnreachable(rc.parent.children, rc.index);
  // Continue executing
  rc.parent.nextChildIndex = rc.index + 1;
  rc.k(rc.val);
}
\end{lstlisting}
\end{minipage}
\end{minipage}
\begin{minipage}{0.5\linewidth}
\begin{lstlisting}[style=smaller]
function execute(node, k, fn, args) {
  node.reachable = true; node.k = k;
  node.index = node.parent.nextChildIndex;
  // Check for input changes
  if (!fnEquiv(node.fn, fn) || !equal(node.args, args)) {
    this.fn = fn; this.args = args;
    // Mark all children as initially unreachable
    markUnreachable(this.children, 0);
    // Call fn with special continuation
    node.nextChildIndex = 0;
    nodeStack.push(node);
    node.entered = true;
    fn(args, function(retval) {
      node = nodeStack.pop();
      // Remove unreachable children
      removeUnreachables(node.children);
      // Terminate early on proposals where
      //    retval does not change
      var rveq = equal(retval, this.retval);
      if (!node.entered && rveq) kexit();
      else {
      	node.entered = false;
      	// retval change may make siblings unreachable
      	if (!rveq)
      	  markUnreachable(node.parent.children,
                          node.index);
      	// Continue executing
      	node.retval = retval;
      	node.parent.nextChildIndex++;
      	k(node.retval);
      }
    });
  } else {
    node.parent.nextChildIndex++;
    k(node.retval);
  }
}
\end{lstlisting}
\end{minipage}

\caption{The main subroutines governing C3's callsite cache. Function calls are wrapped with \ic{cache}, which retrieves (or creates) a cache node for a given address \ic{a}. It calls \ic{execute}, which examines the function call's inputs for changes and runs the call if needed. Finally, MH proposals use \ic{propagate} to resume re-execution of the program from a particular random choice node which has been changed.}
\label{fig:cachingCode}
\end{figure}

When performing an MH proposal, callsite caching aims to avoid re-executing functions and to enable early termination from them as often as possible. In this section, we describe how C3 efficiently implements both of these types of computational `short-circuiting' for probabilistic functional programs. Figure~\ref{fig:cachingCode} provides high-level code for the main subroutines which govern the caching system.

\subsection{Cache Representation}

We first require an efficient cache structure to minimize overhead introduced by performing a cache access on every function call. C3 uses a tree-structured cache: it stores one node for each function call. A node's children correspond to the function's callees. Random choices are stored as leaf nodes.
C3 also maintains a stack of nodes which tracks the program's call stack (\ic{nodeStack} in Figure~\ref{fig:cachingCode}). During cache lookups, the desired node, if it exists, must be a child of the node on the top of this stack. Exploiting this property accelerates lookups, which would otherwise proceed from the cache root. Altogether, this structure provides expected constant time lookups, additions, and deletions. 
In addition, by storing a node's children in execution order, C3 can efficiently determine when child nodes have become `stale' (i.e. unreachable) due to control flow changes and should be removed. A child node is marked unreachable when its parent begins or resumes execution (\ic{execute} line 8; \ic{propagate} line 6) and marked reachable when it is executed (\ic{execute} line 2). Any children left marked unreachable when the parent exits are removed from the cache (\ic{execute} line 16).

\subsection{Short-Circuit On Function Entry}

As described in Section~\ref{sec:compileTime}, every function call is wrapped in a call to \ic{cache}, which retrieves (or creates) a cache node for the current address.
C3 then evaluates whether the node's associated function call must be re-evaluated or if its previous return value can be re-used (the \ic{execute} function). Reuse is possible when the following two criteria are satisfied:
\begin{enumerate}
	\item The function's arguments are equivalent to those from the previous execution.
	\item The \emph{function itself} is equivalent to that from the previous execution.
\end{enumerate}
The first criterion can be verified with conservative equality testing;
C3 uses shallow value equality testing, though deeper equality tests could result in more reuse for structured argument types. Deep equality testing is more expensive, though this can be mitigated using data structure techniques such as hash consing~\cite{HashConsing} or compiler optimizations such as global value numbering~\cite{GlobalValueNumbering}.

The second criterion is necessary because C3 operates on languages with first-class functions, so the identity of the caller at a given callsite is a runtime variable. Checking whether the two functions are exactly equal (i.e. refer to the same closure) is too conservative, however. Instead, C3 leverages the information provided by the function tagging transform from Section~\ref{sec:compileTime}: two functions are equivalent if they have the same lexical ID (i.e. came from the same source location) and if the values of their free variables are equal. C3 applies this check recursively to any function-valued free variables, and it also memoizes the result, as program execution traces often feature many applications of the same function.
This scheme is especially critical to obtain reuse in programs that feature anonymous functions, as those manifest as different closures for each program execution.

\subsection{Short-Circuit On Function Exit}

When C3 re-executes the program after changing a random choice (using the \ic{propagate} function), control may eventually return to a function call whose return value has not changed. In this case, since all subsequent computation will have the same result, C3 can terminate execution early by invoking the exit continuation \ic{kexit}. During function exit, C3's \ic{execute} function detects if control is returning from a proposal by checking if the call is exiting without having first been entered (line 20). This condition signals that the current re-execution originated at some descendant of the exiting call, i.e. \hspace{-2em} a random choice node. 
\begin{wrapfigure}{l}{0.35\textwidth}
\lstset{escapeinside={<@}{@>}}
\begin{lstlisting}
// Using the query table to infer
// the sequence of latent states.
var hmm = function(n, obs) {
  if (n === 0)
    return true;
  else {
    var prev = hmm(n-1, obs);
    var state = transition(prev);
    <@\textcolor{purple}{query.add(n, state);}@>
    observation(state, obs[n]);
    return state;
  }
};

hmm(100, observed_data);
<@\textcolor{purple}{return query;}@>
\end{lstlisting}
\vspace{-1.5em}
\end{wrapfigure}

Early termination is complicated by inference queries whose size depends on model size: for example, the sequence of latent states in an HMM. In lightweight PPL implementations, inference typically computes the marginal distribution on program return values. Thus, a na\"{\i}ve HMM implementation would construct and return a list of latent states. However, this implementation makes early termination impossible, as the list must be recursively reconstructed after a change to any of its elements.

For these scenarios, C3 offers a solution in the form of a global \ic{query} table to which the program can write values of interest. Critically, \ic{query} has a \emph{write-only} interface: since the program cannot read from \ic{query}, a write to it cannot introduce side-effects in subsequent compuation, and thus the semantics of early termination are preserved. Programs that use \ic{query} can then simply return it to infer the marginal distribution over its contents.

\subsection{Optimizations}

C3 takes care to ensure that the amount of work it performs in response to a proposal is only proportional to the amount of the program execution trace affected by that proposal. First, it maintains references to all random choices in a hash table, which provides expected constant time additions, deletions, and random element lookups. This table allows C3 to perform uniform random proposal choice in constant time, rather than the linear time cost of scanning through the entire cache.

Second, proposals may be rejected, which necessitates copying the cache in case its prior state must be restored on rejection. C3 avoids copying the entire cache using a copy-on-write scheme with similar principles to transactional memory~\cite{TransactionalMemory}: modifications to a cache node's properties are staged and only committed if the proposal is accepted. Thus, C3 only copies as much of the cache as is actually visited during proposal re-execution.

Finally, it is not always optimal to cache \emph{every} callsite: caching introduces overhead, and some function calls almost always change on each invocation. C3 detects such callsites and stops caching them in a heuristic process we call \emph{adaptive caching}. A callsite is un-cached if, after at least $N$ proposals, execution has reached it $M$ times without resulting in either short-circuit-on-entry or short-circuit-on-exit. We use $N = 10, M = 50$ for the results presented in this paper. A small, constant overhead remains for un-cached callsites, as calling them still triggers a table lookup to determine their caching status. Future work could explore efficiently re-compiling the program to remove \ic{cache} calls around such callsites.

\section{Experimental Results}
\label{sec:results}

We now investigate the runtime performance characteristics of C3. We compare C3 to Lightweight MH, as well as to systems that use only callsite caching and only continuations. This allows us to investigate the incremental benefit provided by each of C3's components. The source code for all models used in this section is available in the ancillary materials, and our implementation of C3 itself is available as part of the WebPPL probabilistic programming language~\cite{WebPPL}. All timing data was collected on an Intel Core i7-3840QM machine with 16GB RAM running OSX 10.10.2.

\begin{figure}[t!]
\centering
\setlength{\tabcolsep}{2pt}
\begin{tabular}{cc}
    \begin{overpic}[width=0.5\linewidth]{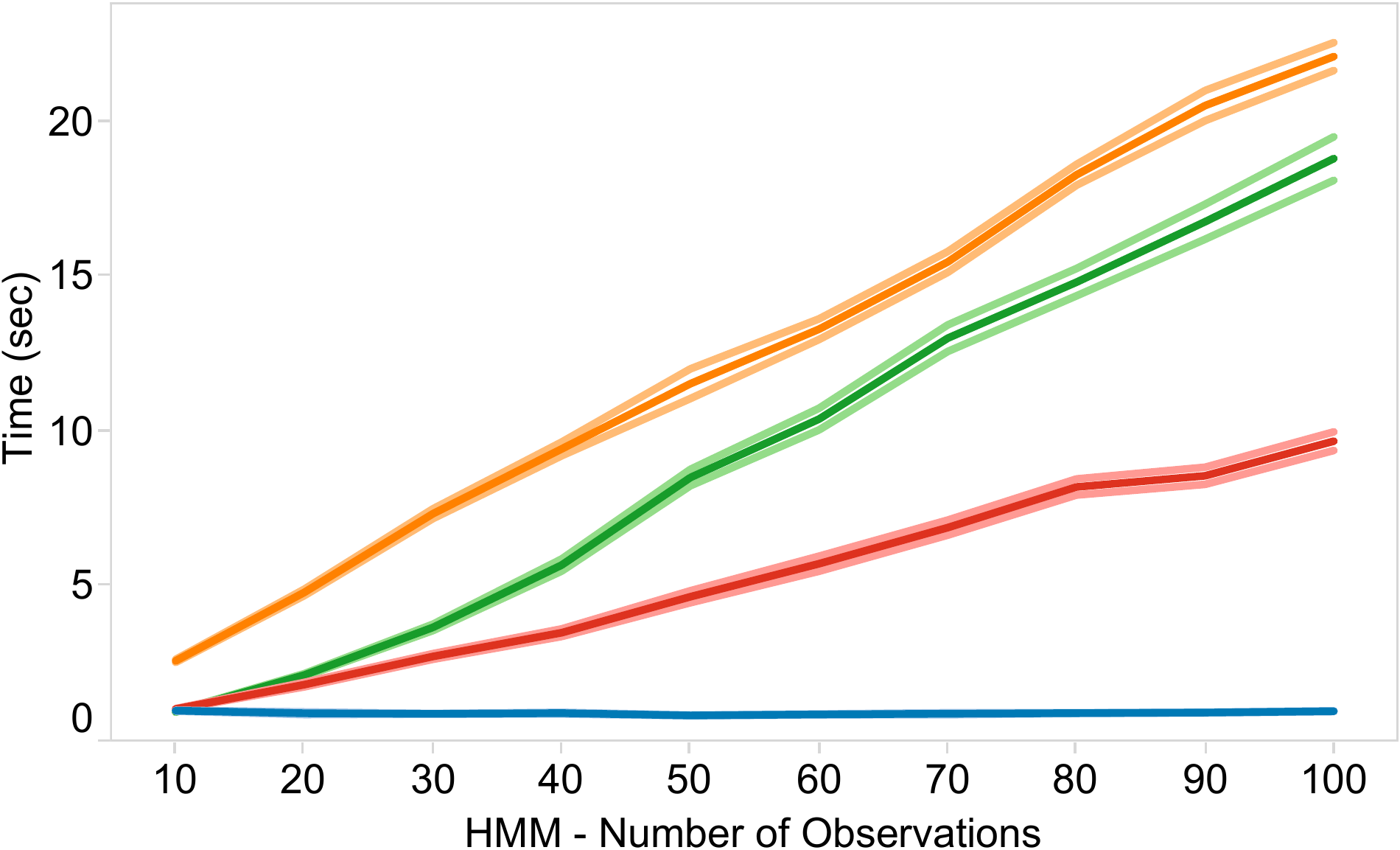}
    	\put(15,37){\includegraphics[width=0.15\linewidth]{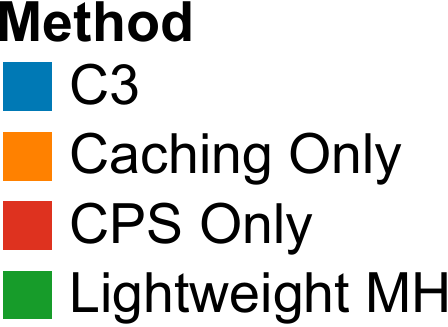}}
    \end{overpic} &
    \includegraphics[width=.5\linewidth]{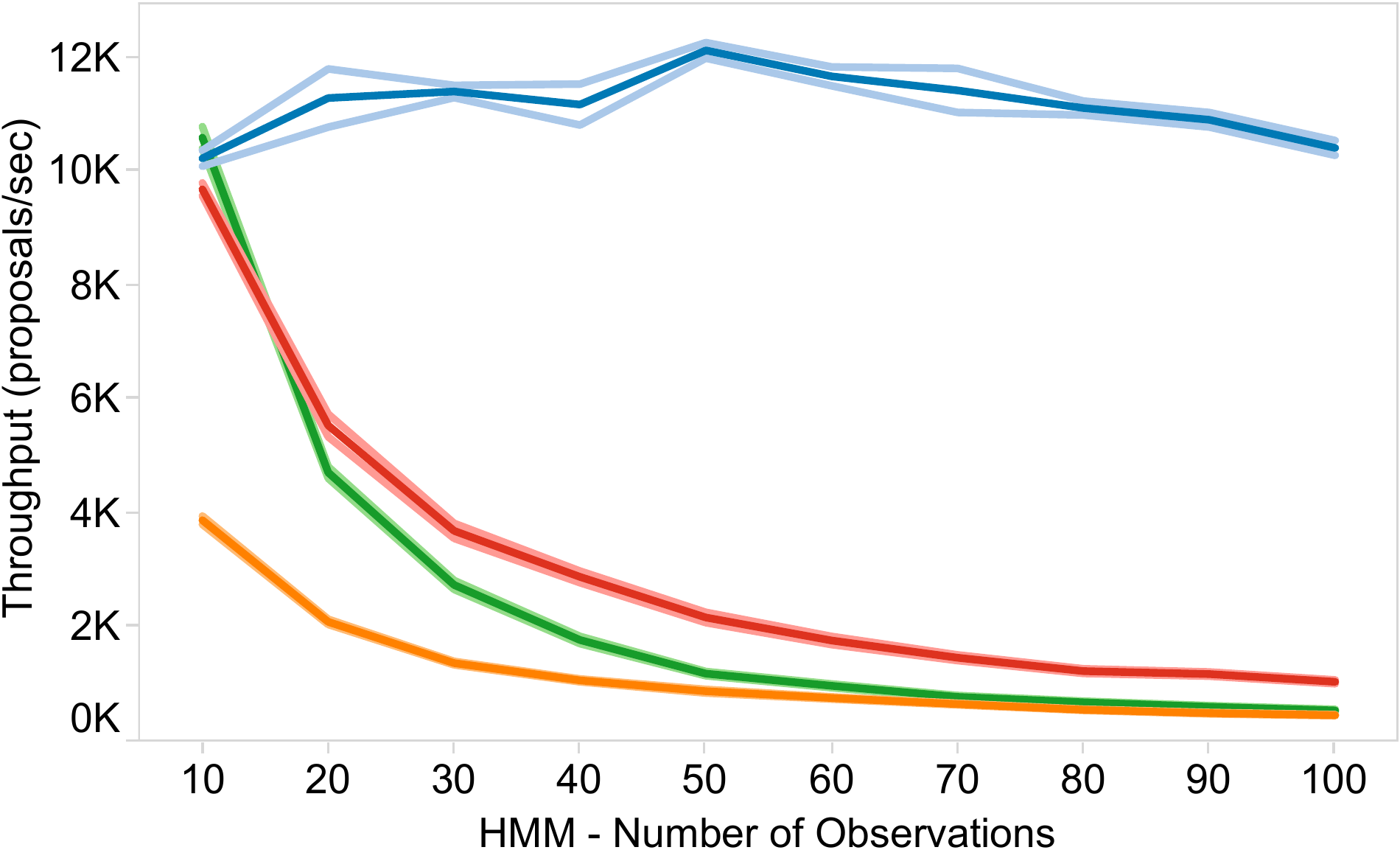}
    \\
    \includegraphics[width=.5\linewidth]{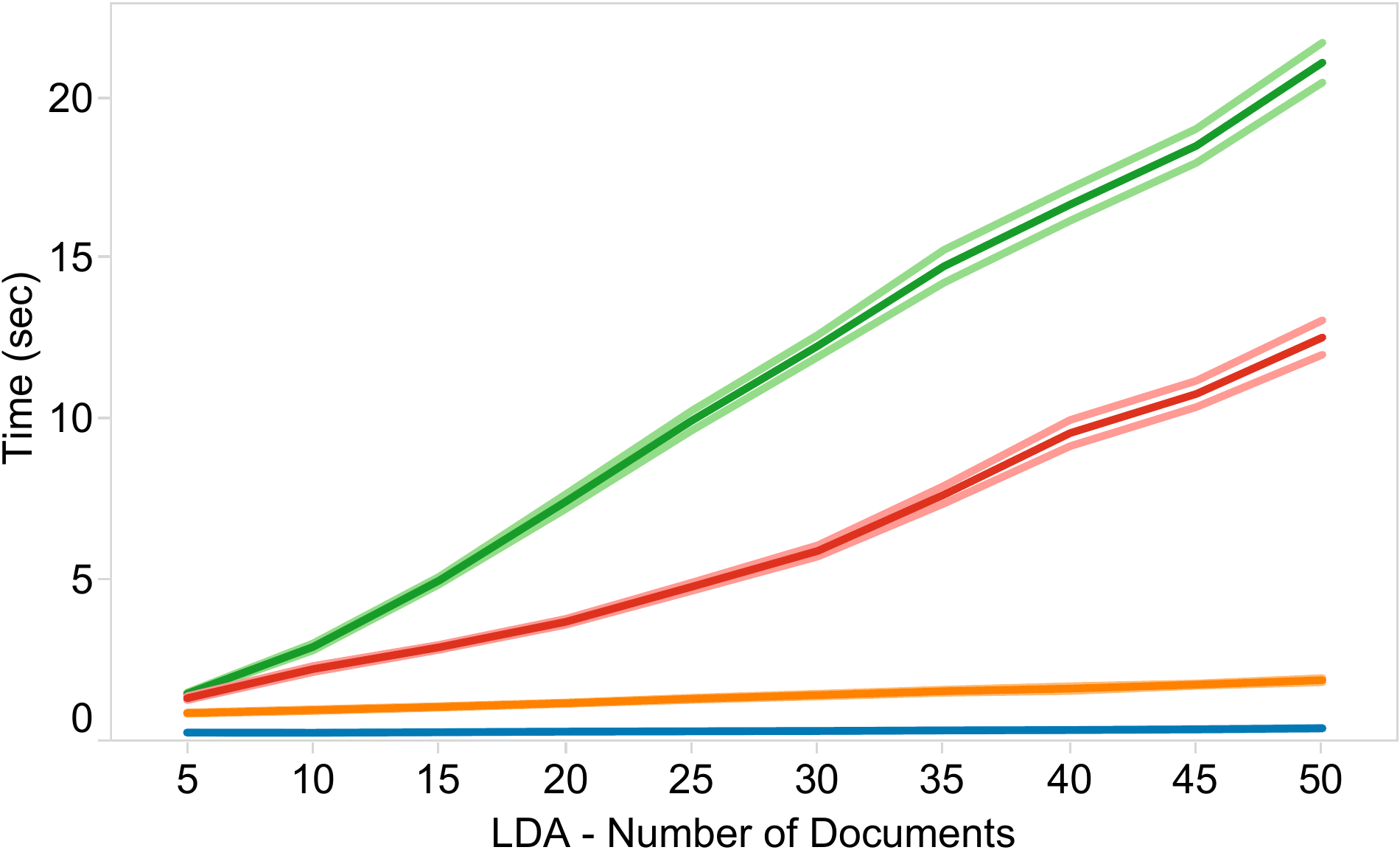} &
    \includegraphics[width=.5\linewidth]{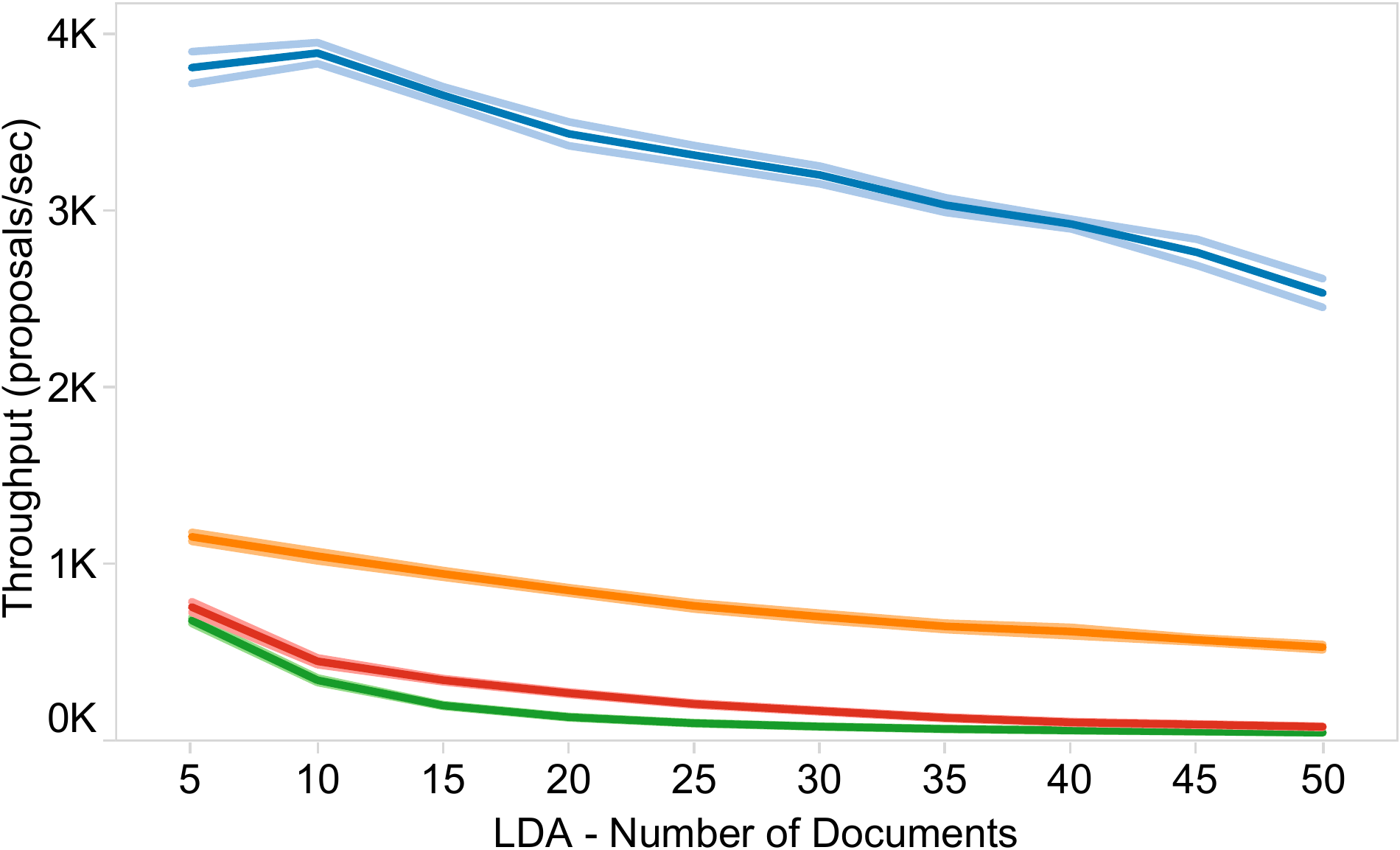}
\end{tabular}
\caption{Comparing the performance of C3 with other MH implementations. \emph{(Top)} Performing 10000 MH iterations on an HMM program. \emph{(Bottom)} Performing 1000 MH iterations on an LDA program. \emph{(Left)} Wall clock time elapsed, in seconds. \emph{(Right)} Sampling throughput, in proposals per second. 95\% confidence bounds are shown in a lighter shade. Only C3 exhibits constant asymptotic complexity for the HMM; other implementations take linear time, exhibiting decreasing throughput.}
\label{fig:hmmAndLda}
\end{figure}

We first evaluate these systems on two standard generative models: a discrete-time Hidden Markov Model and a Latent Dirichlet Allocation model. We use synthetic data, since we are interested purely in the computational efficiency of different implementations of the same statistical inference algorithm.
The HMM program uses 10 discrete latent states and 10 discrete observable states and returns the sequence of latent states. We condition it on a random sequence of observations, of increasing length from 10 to 100, and run each system for 10000 MH iterations, collecting a sample every 10 iterations. The LDA program uses 10 topics, a vocabulary of 100 words, and 20 words per document. It returns the distribution over words for each topic. We condition it on a set of random documents, increasing in size from 5 to 50, and run each system for 1000 MH iterations.

Figure~\ref{fig:hmmAndLda} shows the results of this experiment; all quantities are averaged over 20 runs. We show wall clock time in seconds (left) and throughput in proposals per second (right). For the HMM, C3's runtime is constant regardless of model size, whereas \emph{Lightweight MH} and \emph{CPS Only} exhibit the expected linear runtime (approximately $2N$ and $N$, respectively). As discussed in Section~\ref{sec:approach}, \emph{Caching Only} has the same complexity as \emph{Lightweight MH} but is a constant factor slower due to caching overhead.
For the LDA model, \emph{Lightweight MH} and \emph{CPS Only} all exhibit asymptotic complexity comparable with their performance on the HMM. However, \emph{Caching Only} performs significantly better. The LDA program is structured with nested loops; caching allows re-execution to skip entire inner loops for many proposals. \emph{Caching Only} must still re-execute all ancestors of a changed random choice, though, so it is slower than C3, which jumps directly to the change point. C3 does not achieve exactly constant runtime for LDA because a small percentage of its proposals affect hierarchical variables, requiring more re-execution. This is a characteristic of hierarchical models in general; in this specific case, conjugacy could be leveraged to integrate out higher-level variables.

\begin{figure}[t!]
\centering
\setlength{\tabcolsep}{2pt}
\begin{tabular}{ccc}
	\includegraphics[width=0.25\linewidth]{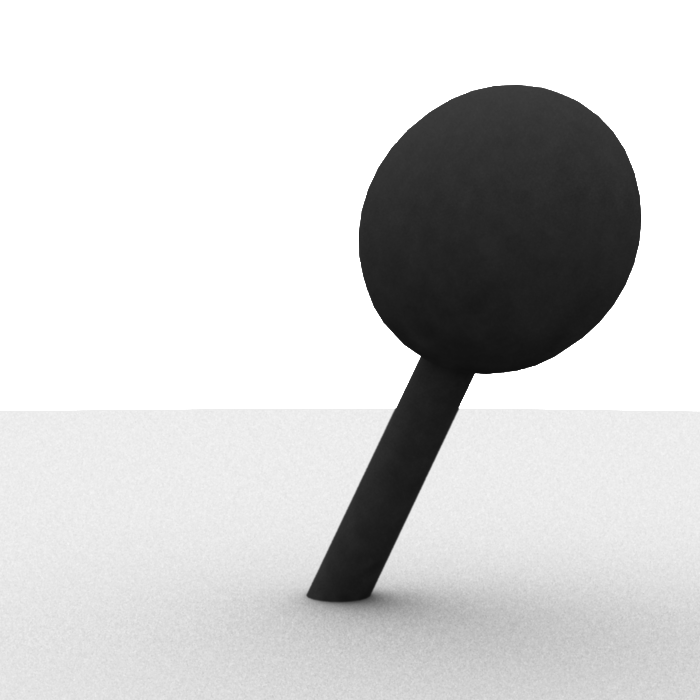} &
	\includegraphics[width=0.25\linewidth]{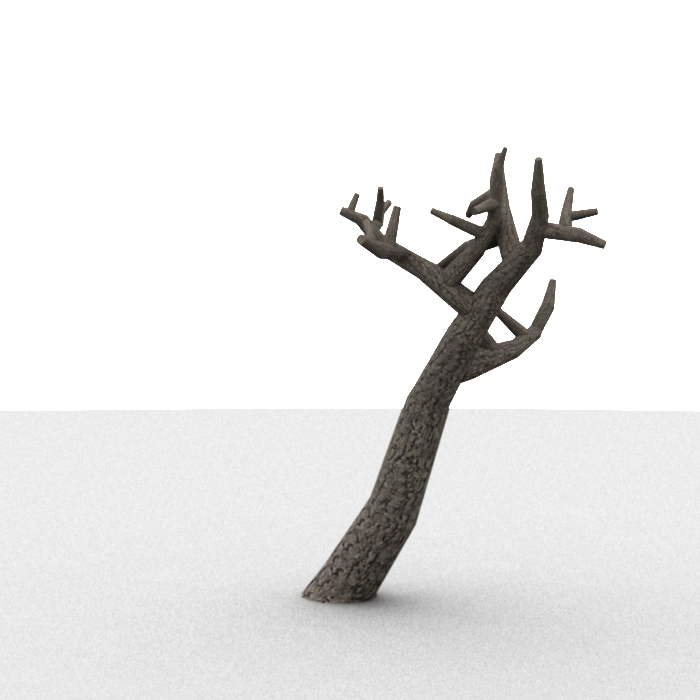} &
	\shortstack{ \includegraphics[width=0.5\linewidth]{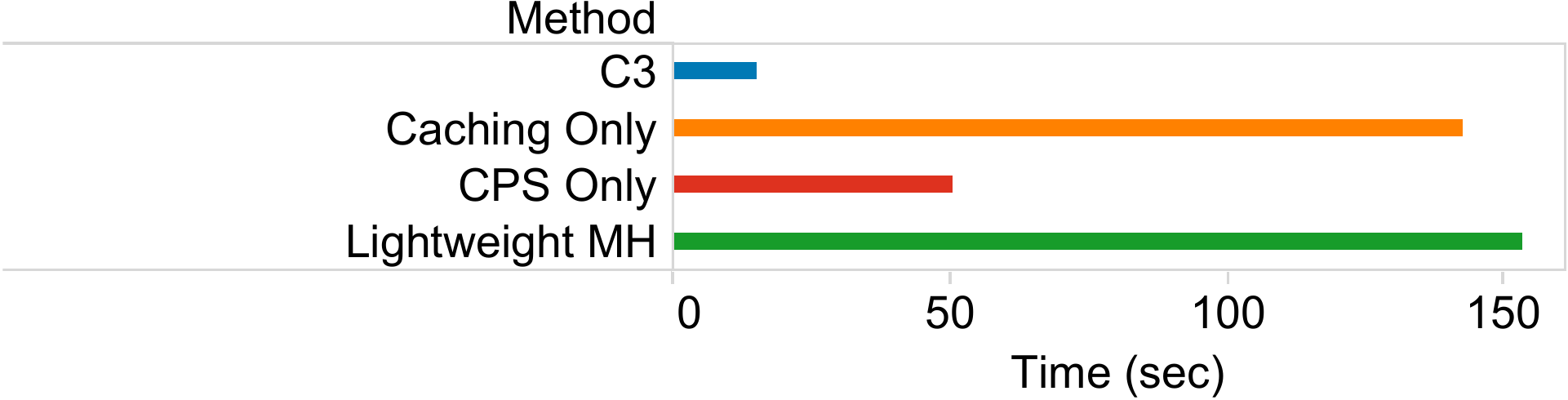} \\ \includegraphics[width=0.5\linewidth]{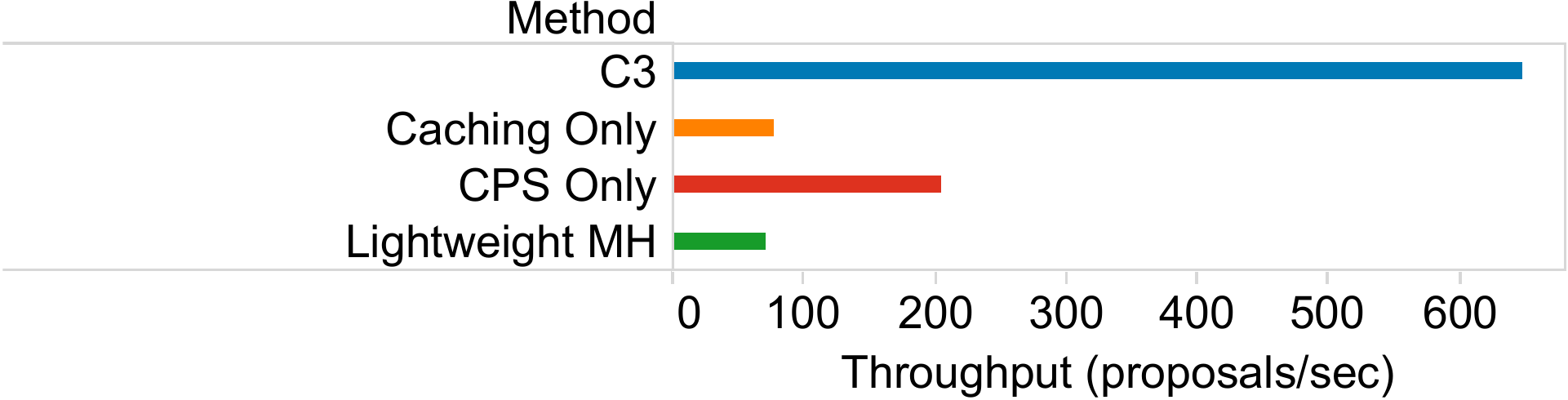} }
\end{tabular}
\caption{Comparing C3 and Lightweight MH on an inverse procedural modeling program. \emph{(Left)} Desired tree shape. \emph{(Middle)} Example output from inference over a tree program given the desired shape. \emph{(Right)} Performance characteristics of different MH implementations. C3 delivers nearly an order of magnitude speedup.}
\vspace{-1em}
\label{fig:procmod}
\end{figure}

We also evaluate these systems on an inverse procedural modeling program. Procedural models are programs that generate random 3D models from the same family. \emph{Inverse} procedural modeling infers executions of such a program that resemble a target output shape~\cite{MPM}. We use a simple grammar-like program for tree skeletons presented in prior work, conditioning its output to be volumetrically similar to a target shape~\cite{SOSMC}. We run each system for 2000 MH iterations.

Figure~\ref{fig:procmod} shows the results of this experiment. C3 achieves the best performance, delivering nearly an order of magnitude speedup over Lightweight MH. Using caching only does not help in this example, since re-executing the program from its beginning reconstructs all of the recursive procedural modeling function's structured inputs, whose equality is not captured by our cache's shallow equality tests.
\begin{wrapfigure}{l}{0.5\textwidth}
\begin{overpic}[width=\linewidth]{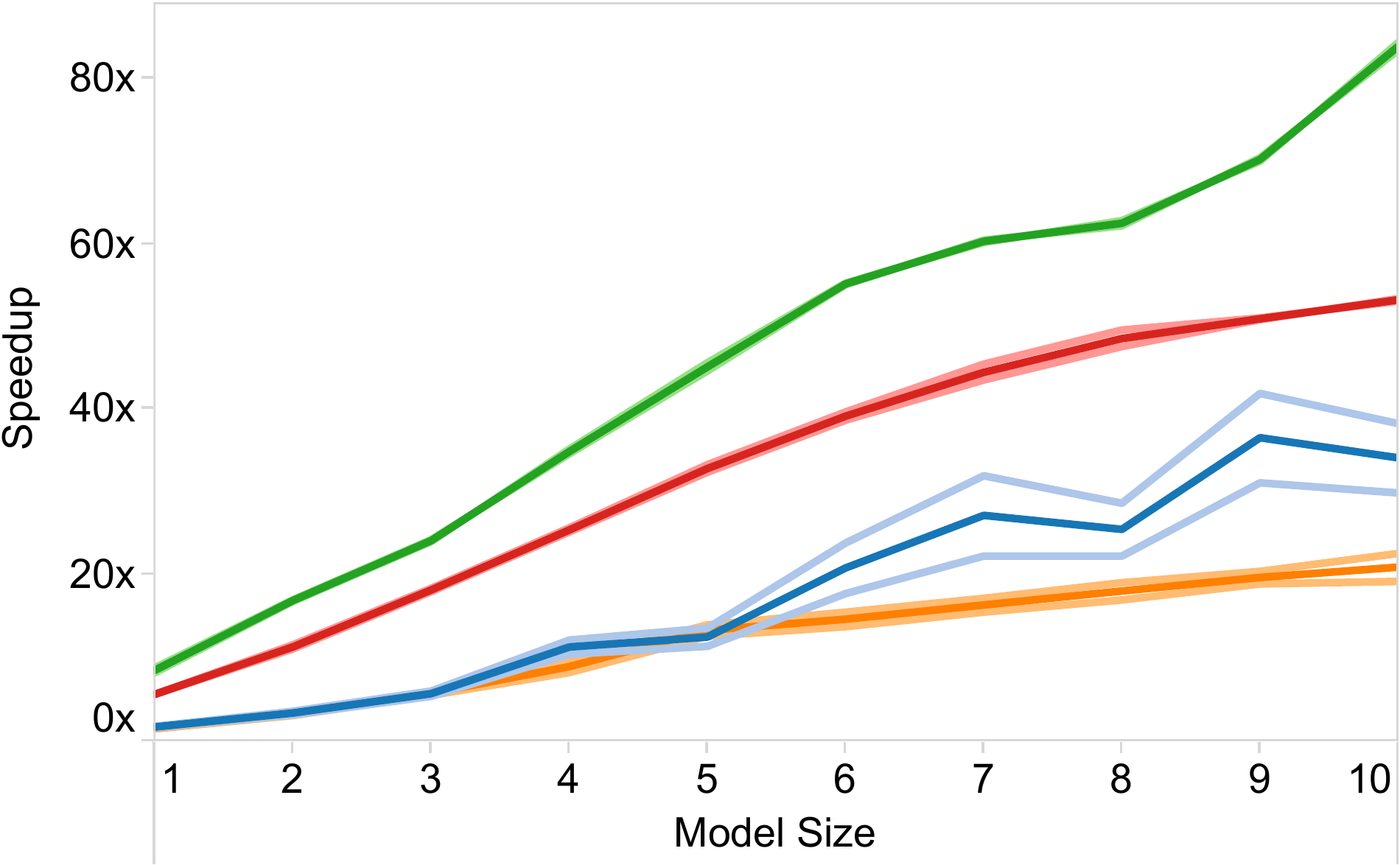}
	\put(15,37){\includegraphics[width=0.14\linewidth]{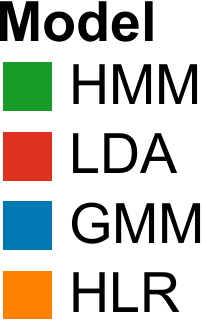}}
\end{overpic}
\vspace{-2em}
\end{wrapfigure}

Finally, the figure on the left shows the results of a wider evaluation: for four models, we plot the speedup obtained by C3 over Lightweight MH (in relative throughput) as model size increases. The four models are: the HMM and LDA models from Figure~\ref{fig:hmmAndLda}, a one-dimensional finite Gaussian mixture model (GMM), and a hierarchical linear regression model (HLR)~\cite{Shred}. The 1-10 normalized Model Size parameter maps to a natural scale parameter for each of the four models; details are available in the ancillary materials. While C3 offers only small benefits over Lightweight MH for small models, it achieves dramatic speedups of 20-100x for large models.

\section{Related Work}
\label{sec:relatedWork}

The ideas behind C3 have connections to other areas of active research. First, incrementalizing MCMC proposals for PPLs falls under the umbrella of \emph{incremental computation}~\cite{IncrementalBiblio}. Much of the active work in this field seeks to build general-purpose languages and compilers to incrementalize any program~\cite{TypeDirectedIncr}. However, there are also systems such as ours which seek simpler solutions to domain-specific incrementalization problems. In particular, C3's callsite caching mechanism was inspired in part by recent work in computer graphics on hierarchical render caches~\cite{IncrementalRendering}.\footnote{An incomplete, undocumented version of C3's callsite caching mechanism also appears in the original MIT-Church implementation of the Church probabilistic programming language~\cite{Church}.}

The Venture PPL features an algorithm to incrementally update a probabilistic execution trace in response to a random choice change~\cite{Venture}. Implemented as part of a custom interpreter, this method walks the trace starting from the changed node, identifying nodes which must be updated or removed, and determining when re-evaluation can stop. C3 performs a similar computation but uses continuations to traverse the execution trace rather than maintaining a complete interpreter state.

The Shred system also incrementalizes MH updates for PPLs~\cite{Shred}. Shred traces a program to remove its control flow and then uses data-flow analysis to produce incremental update procedures for each random choice. This process produces very fast proposal code, but it requires significant implementation cost, and its re-compilation overhead grows very large for programs with high control-flow variability, such as PCFGs. C3's caching scheme is a dynamic analog to Shred's static slicing which does not have compilation overhead but may not be as fast for models with fixed control flow.

The Swift compiler for the BLOG language is another recent system supporting incrementalized MCMC updates~\cite{Swift}. Unlike the above systems, BLOG/Swift uses a \emph{possible-world semantics} for probabilistic programs, representing program state as a graphical model whose structure changes over time. Swift tracks the Markov Blanket of this model, computing incremental updates to it as model structure changes, allowing it to make efficient MCMC proposals. C3 does not explicitly compute Markov blankets, but its short-circuiting facilities limit re-execution to the subset of a changed variable's Markov blanket that is affected by the change.

\section{Discussion and Future Work}
\label{sec:discussion}

This paper presented C3, a lightweight, source-to-source compilation system for incrementalizing MCMC updates in probabilistic programs. We have described how C3's two main components, continuations and callsite caching, allow it both to avoid re-executing function calls and to terminate re-execution early. Our experimental results show that C3 can provide orders-of-magnitude speedups over previous lightweight inference systems on typical generative models. It even enables constant-time updates in some cases where previous systems required linear time. We also demonstrate that C3 improves performance by nearly 10x on a complex, compute-heavy inverse procedural modeling problem. Our implementation of C3 is freely available as part of the open-source WebPPL probabilistic programming language.

Careful optimization of computational efficiency, such as the work presented in this paper, is necessary for PPLs to move out of the domain of research and into production machine learning and AI systems. Along these lines, there are several directions for future work. First, static analysis might allow C3 to determine at compile time dependencies between random choices and subsequent function calls, obviating the need for some input equality checks and reducing caching overhead. Second, C3's CPS transform is overcomplete: it transforms the entire program, but C3 only need continuations at random choice points. Detecting and fusing blocks of purely deterministic code before applying the CPS transform could improve performance. Finally, while the results presented in this paper focus on single-site Metropolis Hastings, C3's core incrementalization scheme also applies to other sampling algorithms, such as Gibbs samplers or particle filter rejuvenation kernels~\cite{Rejuvenation}.

\pagebreak

\bibliographystyle{unsrt}
\bibliography{main}

\begin{thebibliography}{10}

\bibitem{Lightweight}
David Wingate, Andreas Stuhlm\"uller, and Noah~D. Goodman.
\newblock {Lightweight Implementations of Probabilistic Programming Languages
  Via Transformational Compilation}.
\newblock In {\em AISTATS 2011}.

\bibitem{BLOG}
Brian Milch, Bhaskara Marthi, Stuart~J. Russell, David Sontag, Daniel~L. Ong,
  and Andrey Kolobov.
\newblock {BLOG: Probabilistic Models with Unknown Objects}.
\newblock In {\em IJCAI 2005}.

\bibitem{Figaro}
A.~Pfeffer.
\newblock {Figaro: An object-oriented probabilistic programming language}.
\newblock Technical report, Charles River Analytics, 2009.

\bibitem{Church}
Noah~D. Goodman, Vikash~K. Mansinghka, Daniel~M. Roy, Keith Bonawitz, and
  Joshua~B. Tenenbaum.
\newblock {Church: a language for generative models}.
\newblock In {\em UAI 2008}.

\bibitem{Venture}
Vikash~K. Mansinghka, Daniel Selsam, and Yura~N. Perov.
\newblock {Venture: a higher-order probabilistic programming platform with
  programmable inference}.
\newblock {\em CoRR}, 2014.

\bibitem{Anglican}
F.~Wood, J.~W. van~de Meent, and V.~Mansinghka.
\newblock {A New Approach to Probabilistic Programming Inference}.
\newblock In {\em AISTATS 2014}.

\bibitem{Stan}
{Stan Development Team}.
\newblock {\em {Stan Modeling Language Users Guide and Reference Manual,
  Version 2.5.0}}, 2014.

\bibitem{WebPPL}
Noah~D Goodman and Andreas Stuhlm\"{u}ller.
\newblock {The Design and Implementation of Probabilistic Programming
  Languages}.
\newblock \url{http://dippl.org}, 2014.
\newblock Accessed: 2015-5-18.

\bibitem{Quicksand}
Daniel Ritchie.
\newblock {Quicksand: A Lightweight Embedding of Probabilistic Programming for
  Procedural Modeling and Design}.
\newblock In {\em The 3rd NIPS Workshop on Probabilistic Programming}, 2014.

\bibitem{CSI}
Craig Boutilier, Nir Friedman, Moises Goldszmidt, and Daphne Koller.
\newblock {Context-specific Independence in Bayesian Networks}.
\newblock In {\em UAI 1996}.

\bibitem{ContinuationsBook}
Andrew~W. Appel.
\newblock {\em {Compiling with Continuations}}.
\newblock Cambridge University Press, New York, NY, USA, 2007.

\bibitem{HashConsing}
E.~Goto.
\newblock {Monocopy and associative algorithms in an extended lisp}.
\newblock Technical report, 1974.

\bibitem{GlobalValueNumbering}
B.~K. Rosen, M.~N. Wegman, and F.~K. Zadeck.
\newblock {Global Value Numbers and Redundant Computations}.
\newblock In {\em POPL 1988}.

\bibitem{TransactionalMemory}
Maurice Herlihy and J.~Eliot~B. Moss.
\newblock {Transactional Memory: Architectural Support for Lock-free Data
  Structures}.
\newblock In {\em ISCA 1993}.

\bibitem{MPM}
Jerry~O. Talton, Yu~Lou, Steve Lesser, Jared Duke, Radom\'{\i}r M\v{e}ch, and
  Vladlen Koltun.
\newblock {Metropolis Procedural Modeling}.
\newblock {\em ACM Trans. Graph.}, 30(2), 2011.

\bibitem{SOSMC}
Daniel Ritchie, Ben Mildenhall, Noah~D. Goodman, and Pat Hanrahan.
\newblock {Controlling Procedural Modeling Programs with Stochastically-Ordered
  Sequential Monte Carlo}.
\newblock In {\em SIGGRAPH 2015}.

\bibitem{Shred}
Lingfeng Yang, Pat Hanrahan, and Noah~D. Goodman.
\newblock {Generating Efficient MCMC Kernels from Probabilistic Programs}.
\newblock In {\em AISTATS 2014}.

\bibitem{IncrementalBiblio}
G.~Ramalingam and Thomas Reps.
\newblock {A Categorized Bibliography on Incremental Computation}.
\newblock In {\em POPL 1993}.

\bibitem{TypeDirectedIncr}
Yan Chen, Joshua Dunfield, and Umut~A. Acar.
\newblock {Type-Directed Automatic Incrementalization}.
\newblock In {\em PLDI 2012}.

\bibitem{IncrementalRendering}
Michael W\"{o}rister, Harald Steinlechner, Stefan Maierhofer, and Robert~F.
  Tobler.
\newblock {Lazy Incremental Computation for Efficient Scene Graph Rendering}.
\newblock In {\em HPG 2013}.

\bibitem{Swift}
Lei Li, Yi~Wu, and Stuart~J. Russell.
\newblock {SWIFT: Compiled Inference for Probabilistic Programs}.
\newblock Technical report, EECS Department, University of California,
  Berkeley, 2015.

\bibitem{Rejuvenation}
Walter~R. Gilks and Carlo Berzuini.
\newblock {Following a moving target---Monte Carlo inference for dynamic
  Bayesian models}.
\newblock {\em Journal of the Royal Statistical Society: Series B (Statistical
  Methodology)}, 63(1), 2001.

\end{thebibliography}

\end{document}